\documentclass[10pt,twocolumn,letterpaper]{article}

\usepackage{wacv}              
\usepackage{graphicx}
\usepackage{amsmath}
\usepackage{amssymb}
\usepackage{booktabs}
\usepackage{svg}
\usepackage{url}
\usepackage[T1]{fontenc}

\usepackage[pagebackref,breaklinks,colorlinks]{hyperref}

\usepackage[capitalize]{cleveref}
\crefname{section}{Sec.}{Secs.}
\Crefname{section}{Section}{Sections}
\Crefname{table}{Table}{Tables}
\crefname{table}{Tab.}{Tabs.}

\begin{document}

\title{  Generalization of Video-Based Heart Rate Estimation Methods To Low Illumination and Elevated Heart Rates }

\author{Bhargav Acharya, William Saakyan, Barbara Hammer, and Hanna Drimalla\\
{ \textmd Center for Cognitive Interaction Technology (CITEC), Bielefeld University} \\
\textmd Bielefeld, Germany \\
{\tt\small \{bacharya,wsaakyan,bhammer,drimalla\} @techfak.uni-bielefeld.de}
}

\maketitle
\begin{abstract}
\vspace{-1em}
Heart rate is a physiological signal that provides information about an individual's health and affective state. Remote photoplethysmography (rPPG) allows the estimation of this signal from video recordings of a person's face.
Classical rPPG methods make use of signal processing techniques, while recent rPPG methods utilize deep learning networks. 
Methods are typically evaluated on datasets collected in well-lit environments with participants at resting heart rates. 
However, little investigation has been done on how well these methods adapt to variations in illumination and heart rate. 
In this work, we systematically evaluate representative state-of-the-art methods for remote heart rate estimation. Specifically, we evaluate four classical methods and four deep learning-based rPPG estimation methods in terms of their generalization ability to changing scenarios, including low lighting conditions and elevated heart rates.
For a thorough evaluation of existing approaches, we collected a novel dataset called CHILL, which systematically varies heart rate and lighting conditions. The dataset consists of recordings from 45 participants in four different scenarios. The video data was collected under two different lighting conditions (high and low) and normal and elevated heart rates. 
In addition, we selected two public datasets to conduct within- and cross-dataset evaluations of the rPPG methods.
Our experimental results indicate that classical methods are not significantly impacted by low-light conditions. Meanwhile, some deep learning methods were found to be more robust to changes in lighting conditions but encountered challenges in estimating high heart rates.
The cross-dataset evaluation revealed that the selected deep
learning methods underperformed when influencing factors such as elevated heart rates and low lighting conditions were not present in the training
set. 
\end{abstract}

\makeatletter
\renewcommand{\@makefnmark}{}  
\renewcommand{\@makefntext}[1]{\noindent#1}  
\makeatother
\footnotetext{This work was supported by SAIL, funded by the Ministry of Culture and Science of the State of North Rhine-Westphalia under the grant no NW21-059A} 
\section{Introduction}
\label{sec:Intro}

Heart rate (HR) is an important health indicator and its monitoring can help in the early detection of various health problems \cite{cygankiewicz2013heart}.
Furthermore, HR and heart rate variability (HRV) have emerged as valuable tools for predicting and monitoring a person's emotional state \cite{zhou2023dimensional, kumar2023interpretable, 8333392}. 
These biomarkers, HR and HRV, are also influenced by stress and can be used in the prevention of stress-related diseases \cite{wang2018comparative}.

Advancements in signal processing and machine learning methods have given rise to a new class of methods called remote photoplethysmography (rPPG), which directly estimates an individual's heart rate from a video recording of their face \cite{cheng2021deep}. These methods are built on the principles of photoplethysmography (PPG),  a non-invasive technique that uses specialized optical sensors placed on the skin. The methods estimate the HR by measuring changes in reflected light caused by fluctuations in blood volume beneath the skin.

As rPPG methods operate without specialized hardware, videos recorded from mobile phone cameras alone are sufficient for extracting heart rate information\cite{niu2019vipl}. Given the ease with which these methods can be applied, they can be deployed in various real-life scenarios such as telehealth. These real-life scenarios often involve rapid head movements and changes in illumination which have been known to degrade the efficacy of rPPG methods \cite{cheng2021deep}.
Furthermore, hints in the literature suggest that these methods do not generalize well to elevated heart rates \cite{cheng2021deep}, as well as when the videos are compressed. 
Considering the critical nature of the estimated signal, it is essential to rigorously evaluate such methods across diverse conditions that they may encounter in real-world scenarios.

In this work, we target the generalizability of rPPG methods to challenging scenarios, focusing primarily on changes in illumination and elevated heart rates.  
The datasets commonly used for evaluating rPPG estimation approaches typically contain little to no variation in illumination and only collect the resting heart rates \cite{bobbia2019unsupervised, Heusch_ARXIV_2017, stricker2014non}.
As a first contribution of this article, we introduce the CHILL dataset, a novel dataset specifically designed to incorporate challenging conditions such as low illumination and elevated heart rates. 
We leverage this dataset to conduct systematic evaluations of commonly used rPPG methods, encompassing both deep learning and computer vision-based approaches.

In summary, the main contributions of this work are as follows:
\begin{enumerate}
    \item we collected a novel dataset, CHILL, consisting of 45 participants recorded under four different scenarios, which include high HR and low illumination. We make this dataset available to other researchers.
    \item we systematically evaluate commonly used rPPG estimation methods, specifically four classical methods and four deep learning (DL) based methods. We carry out our evaluations on two publicly available datasets, COHFACE \cite{Heusch_ARXIV_2017} and PURE \cite{stricker2014non}, and our collected CHILL dataset. 
    \item We investigate the generalizability of DL-based rPPG methods through cross-dataset evaluations.
\end{enumerate}
\section{Related Work}

In this section, we provide an overview of the existing work on heart rate estimation methods. We discuss two main categories of methods: classical and deep learning-based.
In addition, we describe the challenges these methods face to generalize under various real-life scenarios, such as low illumination and elevated heart rate.

\subsection{Classical Methods}
Early work demonstrated that rPPG estimation was possible using consumer-grade cameras and ambient light \cite{verkruysse2008remote}. One of such early methods, GREEN\cite{verkruysse2008remote}, showed that it was possible to use the green channel of the RBG video to extract the rPPG signal, as hemoglobin absorbs more green light compared to red and blue, and in turn, estimate the HR.

Subsequent works incorporated the knowledge of how light interacts with the skin into the methods. This interaction of light was first modeled by Wang et al. \cite{wang2016algorithmic}, who introduced the skin reflection model. This model considers that the light captured by the camera, which is reflected from the skin, consists mainly of two components: specular and diffuse. 
Specular reflection is the surface-level skin reflection and does not contain relevant HR information. In contrast, diffuse reflection corresponds to light that is reflected from the skin tissue and blood vessels, which contain information on the changing blood volume. 
Methods such as CHROM \cite{de2013robust} and POS \cite{wang2016algorithmic} were developed to eliminate these extraneous specular reflections. CHROM \cite{de2013robust} does this by considering the differences in the color channels, while POS \cite{wang2016algorithmic} uses a projection of the reflected light onto a plane orthogonal to the skin. 
Other works use statistical dimensionality reduction techniques such as PCA \cite{6078233} and ICA \cite{poh2010advancements} to estimate the rPPG signal.

\begin{table*}
    \centering 
    \caption{rPPG datasets}
    \label{tab: public Datasets}
    \begin{tabular}{cccccc}
        \toprule 
        \toprule 
        \textbf{Dataset} & \textbf{Participants} &  \textbf{Lighting conditions} & \textbf{FPS} & \textbf{Resolution } & \textbf{HR range}\\ 
        \midrule
        \toprule 
        COHFACE \cite{Heusch_ARXIV_2017}& 40(F:12, M:28) & Natural and studio & 20 & 640x480 & 45-97\\ 
        PURE \cite{stricker2014non} & 10(F:2, M:8) & Natural & 30 & 640x480 & 42-148 \\ 
        \textit{CHILL } & 45(F:27, M:17) & Studio (bright and dark)& 25 & 1920x1080 & 54-141 \\ 
    \bottomrule
    \end{tabular}
\end{table*}
\subsection{Deep Learning Based Methods}

The field of computer vision has witnessed a surge in deep learning methods, leading to their growing prevalence over classical approaches. This trend extends to rPPG estimation, where numerous deep learning-based methods have emerged, offering significant advantages.
These deep learning methods can be broadly classified into two main categories end-to-end methods and hybrid methods \cite{cheng2021deep}. End-to-end methods consist of a deep learning architecture that can directly process video frames and output the rPPG signal. 
Hybrid methods use deep learning architectures within their pipeline along with other signal processing methods, where the deep learning architectures are used for different tasks ranging from signal optimization to signal extraction.

The initial methods introduced for rPPG estimation predominantly involved end-to-end deep learning approaches.
Spetlik et al. \cite{vspetlik2018visual} were one of the first to show that end-to-end deep learning based approaches can be used for the task of rPPG estimation, utilizing 2D-CNNs within their architecture. In addition, they made use of a second network called the extractor to estimate the HR from the predicted rPPG signal. 
Yu et al. \cite{yu2019remote} experimented with methods that incorporated temporal dimensions of the video input. This led to the development of Physnet \cite{yu2019remote}, which used 3D-CNNs instead of 2D-CNNs. They also experimented with the use of LSTMs which performs worse compared to Physnet \cite{yu2019remote}.

Several recent methods have been developed to address specific challenges in rPPG estimation, including motion artifacts and video compression. One such method, DeepPhys by Cheng et al. \cite{chen2018deepphys}, specifically targets motion artifacts.
The DeepPhys architecture consists of two branches, one for motion and one for appearance. Each branch consists of 2D-CNNs based on the VGG architecture \cite{simonyan2014very}.
These two independent branches are connected by an attention module, which directs the model to focus on relevant areas of the image corresponding to the rPPG signal. 
Liu et al. \cite{liu2021multitask}, similar to DeepPhys \cite{chen2018deepphys}, proposed a two-branch architecture that could estimate the respiration rate along with the heart rate. They utilized temporal shift convolutions \cite{lin2019tsm}, which helped reduce the number of training parameters without sacrificing temporal information.
These smaller models could be deployed on mobile platforms with limited processing power. 

Yu et al. proposed STVEN \cite{yu2019compress} and rPPGnet \cite{yu2019compress} to mitigate the loss in performance of rPPG methods on highly compressed videos. The STVEN architecture enhances the highly compressed videos which are then processed by rPPGnet to estimate the rPPG signal. 
The rPPGnet \cite{yu2019compress} utilizes a spatiotemporal convolutional network, which takes in 64 consecutive frames of the input video and outputs the corresponding rPPG signal. Additionally, the model incorporates a skin detection-based attention module to eliminate the influences of non-skin regions of the video.

In recent works, vision Transformers \cite{dosovitskiy2021image}, originally utilized for processing video data in tasks such as action recognition, video inpainting, and 3D animations \cite{han2022survey}, have also been applied to the task of rPPG estimation \cite{kang2022transppg, liu2021efficientphys, yu2022physformer}. This includes methods such as TransPPG\cite{kang2022transppg}, EfficientPhys \cite{liu2021efficientphys}, and PhysFormer\cite{yu2022physformer}.

\subsection{Generalization of Methods}
\subsubsection{Low Light Conditions}
Most rPPG methods are based on the skin reflectance model \cite{wang2016algorithmic} and are highly dependent on the amount of ambient light present. Due to this, it can be challenging to accurately estimate the rPPG signal when the skin is not well illuminated. However, most of the publicly available datasets are recorded in well-lit environments. These environments are illuminated either by artificial light sources or natural light in a controlled manner \cite{niu2019vipl, Heusch_ARXIV_2017, stricker2014non}. 
Datasets from such controlled environments are often used to train and evaluate rPPG methods \cite{liu2021multitask, chen2018deepphys, yu2019remote}.
Such methods, when not rigorously tested, could lead to misleading predictions in real-life situations where the environment is less controlled. Yang et al. \cite{yang2022assessment} attempted to address this issue by collecting a dataset with illumination variance. The dataset consists of multiple scenarios in which the intensity of light on the participant's face is varied. 
This setting is important to see how the methods adapt when only certain parts of the face are illuminated. However, there is a need for scenarios where the overall illumination is reduced to evaluate the methods when there is a lack of light reflecting from the individual's skin. The apparent lack of research on how adversely the methods or models are affected when there is a drop in illumination is of concern.

\subsubsection{Elevated Heart Rates}

In a recent review \cite{cheng2021deep} on deep learning-based heart rate estimation algorithms, Cheng et al. point out the lack of research on how elevated heart rate affects the performance of deep learning-based methods. Cheng et al. emphasize that in  RePSS 2020 \cite{li20201st}, the first challenge on remote physiological signal sensing, the top three models performed better when the heart rate was between 77 and 90 bpm and worse when it was above 90 bpm. 
Li et al. \cite{8373836} attempted to tackle this issue by collecting a dataset named OBF. This dataset consists of videos of the face with ground truth PPG of participants pre- and post-exercise. However, this dataset is currently not publicly available. Available Datasets such as COHFACE \cite{Heusch_ARXIV_2017} and PURE \cite{stricker2014non} record participants with a resting heart rate. VIPL \cite{niu2019vipl} has a setting in which the video is recorded post-exercise but is collected in a well-lit environment. 

The current reliance on controlled datasets creates a critical knowledge gap in how rPPG methods perform and transfer their capabilities across diverse real-world conditions, encompassing both high heart rates and low illumination. 
To address this limitation, a dataset specifically designed to represent these challenging scenarios is necessary for systematic evaluation of rPPG methods.
\section{Methods and Material}

In this section, we present the datasets that are used for the evaluation and the methods that will be evaluated. We introduce the two public datasets and describe the experimental setup for collecting our novel dataset. 
\autoref{tab: public Datasets} summarizes the key characteristics of all datasets, including lighting conditions, FPS, resolutions, and the range of recorded heart rates.
Finally, we outline the selected classical methods and DL-based methods.

\subsection{Public Datasets}
\label{subsec: public datasets}
We utilised two public datasets, namely COHFACE\cite{Heusch_ARXIV_2017}, and PURE \cite{stricker2014non}, for the evaluations. 
These datasets were specifically chosen because they were collected in controlled laboratory environments with minimal variations in illumination and heart rates.

COHFACE \cite{Heusch_ARXIV_2017} consists of 40 participants, each recorded with a digital camera at a frame rate of 20 Hz. The ground truth pulse was simultaneously recorded with a contact device at a sampling rate of 256 Hz. Each participant was recorded twice in two different scenarios, for a total of four videos per participant. The two different scenarios consisted of two different lighting conditions: good and natural. The good condition used a halogen spotlight with additional ceiling lights. The natural condition used natural light coming through the window.

The PURE dataset \cite{stricker2014non} consists of 10 participants. Each participant was recorded under natural lighting conditions using an eco274CVGE camera at a frame rate of 30 Hz. The ground truth was simultaneously measured with a finger-clip pulse oximeter at a sampling rate of 60 Hz. The setup consisted of placing a participant at a distance of 1.1 meters from the camera and recording during the daytime. Natural light through a frontal window was the only light source that was used to illuminate the environment. The authors also point out that there was a change in illumination due to moving clouds. The recordings consisted of six scenarios per participant (i.e., steady, talking, slow translation, fast translation, small rotation, and medium rotation).

\subsection{CHILL Dataset}
\label{subsec: CHILL dataset}

To address the gap in publicly available datasets lacking low-light conditions and elevated heart rates, we collected a novel dataset called CHILL (\textbf{C}hallenging \textbf{H}eartrate and \textbf{Il}lumination). This dataset consists of synchronized video recordings of individuals' faces and their corresponding ground truth PPG signals. The recordings were captured under varied lighting conditions, with participants exercising to induce high and low heart rates. In this section, we describe the experimental design and recording setup used to collect the CHILL dataset.

\subsubsection{Data collection procedure}
\begin{figure*}
    \centering
    \includegraphics[width=\linewidth]{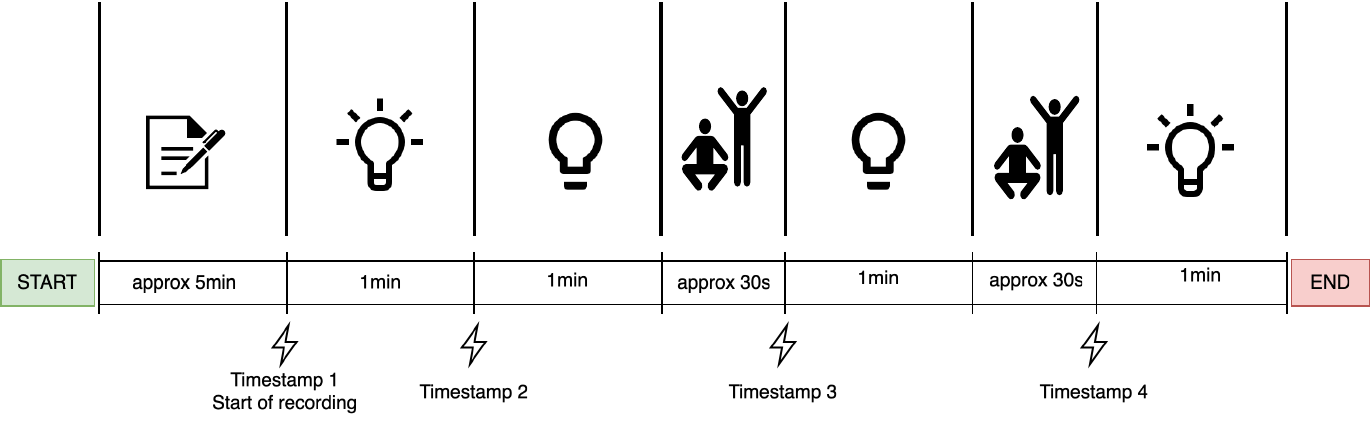}
    \caption{Data collection protocol}
    \label{fig:data collection protocol}
\end{figure*}

The collected dataset consists of four video recordings of each participant's face and a time-aligned ground truth PPG sensor signal. The data collection process is illustrated in \autoref{fig:data collection protocol} and consists of 4 different recording scenarios of 1 minute each. The study was approved by the local ethics committee. 

\begin{figure}[!ht]
    \centering
    \includegraphics[trim={0 2.5cm 0 2cm},clip,width=\linewidth]{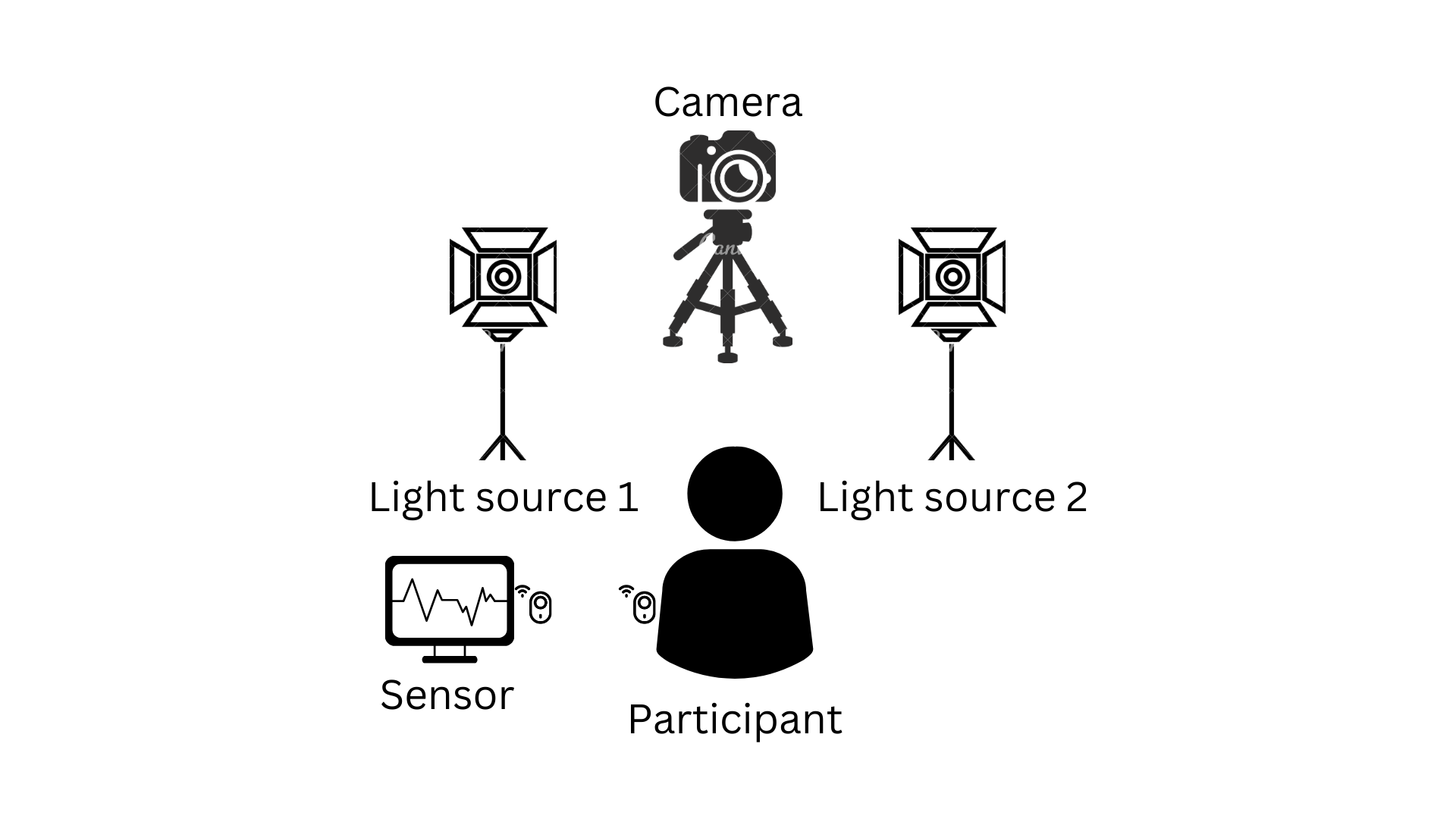}
    \caption{Data collection setup}
    \label{fig: data collection setup}
\end{figure}

Participants were recruited through flyers advertising the study at the university.
All recordings took place in a university laboratory. 
After participants gave their informed consent, the experimenter placed the PPG and electrocardiogram (ECG) sensors on the participant. The clip-on PPG sensor was placed on the left index finger of the participant. 
The ECG sensor consisted of three electrodes, two placed below the collarbone on opposite sides, and the third electrode positioned near the lower right rib cage.
The participants were then led into a room with an experimental setup as depicted in \autoref{fig: data collection setup} and asked to sit still facing the camera. 
The windows in the recording room were covered with tight shutters to block out external light. The only light sources illuminating the environment were two LED array light sources. These light sources were both set to their maximum power setting (indicated as 50 on the device). All the recordings took place only when the participant was seated facing the camera.  
In setting 1 (LowHR-Bright), recordings consisted of participants in a bright environment with normal heart rates. The illumination for the second setting (LowHR-Dark) was changed by adjusting the output of both the light sources to 5 (output of the device range from 0-50), resulting in a dark environment. For the last two settings, the participants were asked to perform short exercises (pushups or squats) before sitting still in front of the camera again. This resulted in settings where the participants had elevated heart rates. For setting 3 (HighHR-Dark), the lighting was similar to that of setting 2. For setting 4 (HighHR-Bright), the illumination was increased, similar to setting 1. 
To maintain a consistent distribution of high heart rates across varying lighting scenarios, we employ random shuffling of the lighting order. This process yields two distinct orders for the study.

\subsubsection{Recording Setup}
The videos of participants’ faces were recorded using a DSLR camera (CanonEOS 550D). The Biosignalplux explorer kit was used to collect the ground truth PPG and ECG at the same time. The videos were recorded at a resolution of 1920x1080. The frame rate was 25 fps, and the ground truth was sampled at 1000 Hz. Timestamps recorded within the sensor software at regular intervals were used to achieve synchronization between the video recording and the sensor data.

\subsubsection{Publication of the Dataset}
The participants were provided with a consent form that asked them for the publication of their dataset.
The collected data is made available with anonymization. The anonymization consists of downsampling each frame in the video to $128 \times 128$ pixels, as used in the experiments. The data of all the participants who have consented to share their data is available online through zenodo.org at \url{https://doi.org/10.5281/zenodo.14637544}. 

\subsection{Classical Methods}
We selected four classical methods, which are based on signal processing techniques, for our evaluations. The chosen methods are GREEN \cite{verkruysse2008remote}, POS \cite{wang2016algorithmic}, CHROM \cite{de2013robust}, and ICA \cite{poh2010advancements}, as they are commonly used as benchmark methods for rPPG estimation.

rPPG-ToolBox \cite{liu2022rppg}, an open-source Python framework, was used to implement the classical methods. 
The toolbox additionally provides face tracking and cropping algorithms that are generally used to pre-process the videos. 
The toolbox was used to preprocess and extract the mean RGB signal from the input videos. 
The selected rPPG estimation methods, ICA \cite{poh2010advancements}, CHROM\cite{de2013robust}, GREEN\cite{verkruysse2008remote}, and POS\cite{wang2016algorithmic}, were used to extract a rPPG signal from the RGB signal. 

\subsection{DL Methods}
In this work, we focus on evaluating end-to-end DL methods. We select four methods that have been prominently used for the task of rPPG estimation, namely, DeepPhys \cite{chen2018deepphys}, TS-CAN\cite{liu2021multitask}, Physnet\cite{yu2019remote}, and rPPGNet \cite{yu2019compress}.
An overview of all considered deep learning methods is provided in \autoref{tab: a1}. 
The table also includes the datasets used by the original authors to train their respective models.

\begin{table*}
    \centering 
    \caption{Overview of the DL methods}
    \begin{tabular}{ccccc}
        \toprule 
        \toprule 
        \textbf{Methods} & \textbf{Network} & \textbf{Training Datasets} & \textbf{Face detector} & \textbf{lr} \\ 
        \midrule
        \toprule 
        DeepPhys \cite{chen2018deepphys}   & 2D-CNN          & Private Dataset                & Viola-Jones \cite{990517}& 1.0 \\
        TS-CAN  \cite{liu2021multitask}    & TS-CNN          & AFRL \cite{6974121}                           & No  & 1.0 \\
        Physnet \cite{yu2019remote}        & 3D-CNN          & OBF \cite{Li2018TheOD}                           & Viola-Jones \cite{990517} & 1e-4 \\ 
        rPPGNet  \cite{yu2019compress}     & 3D-CNN          & OBF \cite{Li2018TheOD} and MAHNOB-HCI \cite{5975141}.            & Viola-Jones \cite{990517} & 1e-4 \\ 
    \bottomrule
    \label{tab: a1}
    \end{tabular}
\end{table*}

\subsubsection{Preprocessing for DL  methods} 

All raw videos are preprocessed before they are passed to the DL methods. The preprocessing is dependent on the DL method that is considered.
For DeepPhys \cite{chen2018deepphys} and TS-CAN \cite{liu2021multitask}, the preprocessing for the appearance branch consisted of downsampling each frame to  $ 36 \times 36 $ pixels. For the motion branch, the inputs were normalized using adjacent frames. The normalization was performed as follows, where $c(t)$ represents a frame at time $t$ :

\begin{equation}
    \frac{c(t+1) -c(t)}{c(t) + c(t+1)}
\end{equation}

For Physnet \cite{yu2019remote} 
and rPPGnet \cite{yu2019compress} the preprocessing involved cropping raw frames using the Viola-Jones face detector \cite{990517}. Subsequently, the cropped faces were resized to 128x128.
The rPPGnet \cite{yu2019compress} uses an additional binary skin mask as an input along with the raw frames. These skin maps were generated using the open source package, Bob \footnote{https://gitlab.idiap.ch/bob/bob.ip.skincolorfilter}, with a threshold of 0.3.

\subsubsection{Training configurations} All the DL models were trained on NVIDIA A40 GPUs. 
The rPPG-Toolbox \cite{liu2022rppg} was employed for TS-CAN \cite{liu2021multitask}, DeepPhys \cite{chen2018deepphys}, and Physnet \cite{yu2019remote}. 
For rPPGnet\cite{yu2019compress},
the implementation provided by the original authors was used used to extend the toolbox. 

The training process was consistent across datasets, with no changes to optimizers or pipelines. Batch sizes were adjusted based on GPU memory constraints.

The loss functions varied across models: DeepPhys \cite{chen2018deepphys} and TS-CAN \cite{liu2021multitask} employed mean squared error (MSE), while Physnet \cite{yu2019remote} and rPPGnet \cite{yu2019compress} used negative Pearson correlation. Notably, rPPGnet \cite{yu2019compress} incorporated binary cross-entropy loss for its skin segmentation module.
For a more comprehensive understanding of these loss functions, we recommend consulting the original papers.

\subsection{HR estimation} 
The estimated rPPG signal, from classical and deep learning methods, was passed through a bandpass filter and the filtered signal was used to calculate the heart rate.
The HR was obtained by estimating the power spectral density (PSD) of the rPPG signal. 
The rPPG-Toolbox \cite{liu2022rppg} by default resorts to using a periodogram to estimate the PSD. However, we opt for using Welch's method \cite{1161901}, an improvement over the standard periodogram that reduces noise in the estimated PSD. The frequency corresponding to the maximum density is considered the estimated heart rate.

\subsection{Evaluation Metric}
We used mean absolute error (MAE) as the evaluation metric, which is expressed using the following formula:
\begin{equation}
    MAE = \frac{1}{T} \sum_{i=1}^{T} | HR_{GT} - HR_{EST}| 
\end{equation}
where $HR_{GT}$ is the ground truth heart rate and $HR_{EST}$ is the estimated heart rate. $T$, refers to the total number of videos evaluated.
A dummy estimator that predicts the mean, which is calculated using the training set, was used as an indicator to see how well the models performed.

\section{Experiments and Results}

To systematically evaluate the different methods, we conduct experiments on the datasets that were described in \ref{subsec: public datasets} and \ref{subsec: CHILL dataset}.
In this section, we present the novel dataset that was collected using the protocol described in \ref{subsec: CHILL dataset}. We further describe the experiments conducted and provide an overview of the results. 

\subsection{Collected Dataset}

\begin{figure}
    \centering
    \includegraphics[width=0.9\linewidth]{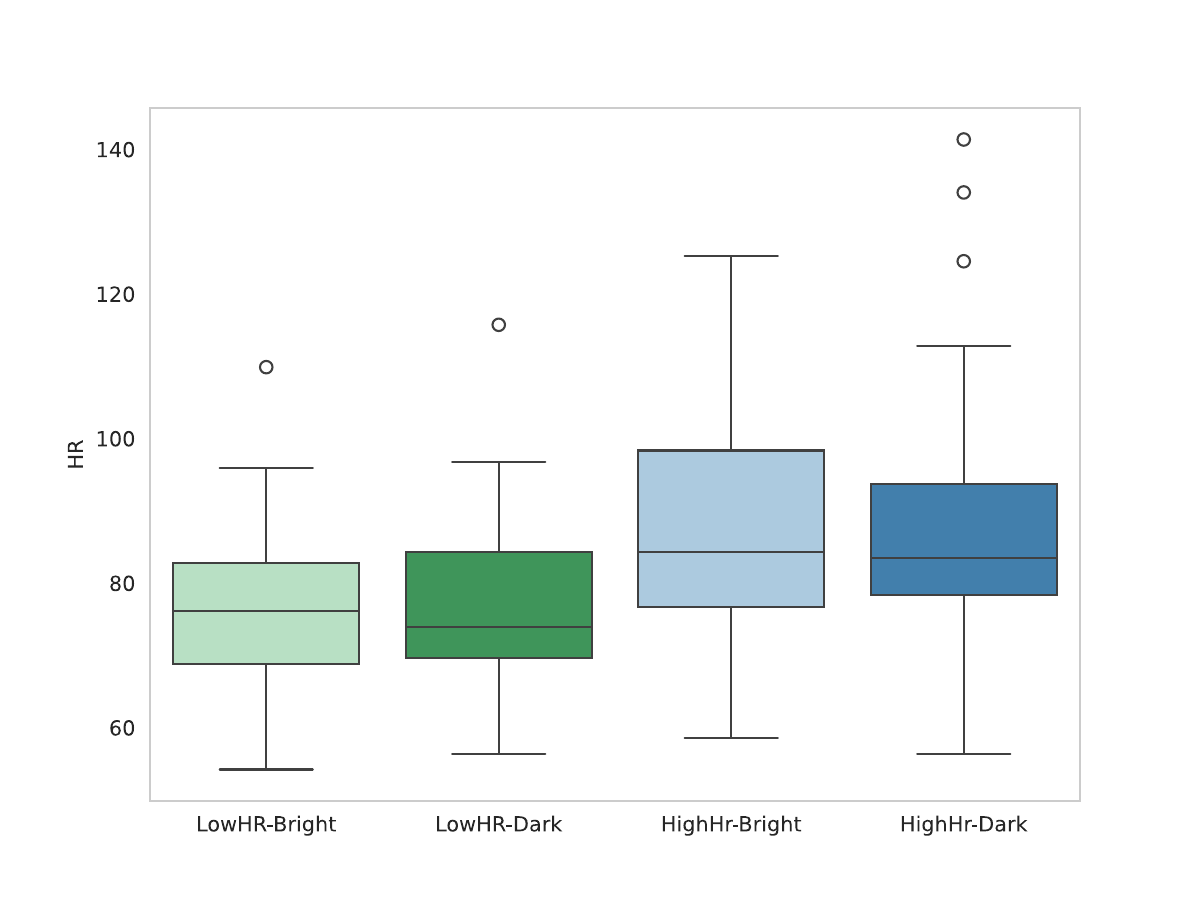}
    \caption{Participants' HR per setting for CHILL dataset}
    \label{fig: box plot scenarios CHILL}
\end{figure}

We collected data from 50 participants. However, due to missing ground truth data caused by faulty sensors, 5 participants were excluded. This resulted in a final dataset containing video recordings of 45 participants.
The ground truth HR was estimated using the collected ground truth PPG signal, using Welch's method \cite{1161901} to estimate the PSD.
The spread of the heart rate per setting is depicted in \autoref{fig: box plot scenarios CHILL}. The HR of the participants ranges from 54 to 141 beats per minute. The mean of the HR for LowHR and HighHR settings were 76.2 and 87.3 respectively. The variance of heart rate in setting HighHR-Bright is higher compared to HighHR-Dark, which can be seen in \autoref{fig: box plot scenarios CHILL}.
The average pixel values for dark and bright settings were 33.6 and 129.7 respectively.  
The skin types of the recorded participants consisted of type 1, 2, and 3 on the Fitzpatrick scale (ranging from 1 to 6). 

The dataset shows that the exercises that preceded scenarios 3 and 4 do result in higher heart rates compared to scenarios 1 and 2. 

\subsection{Evaluation of Methods on all Datasets}
\label{subsec: training on datasets}

We conducted a systematic evaluation of all selected models across all datasets. For the deep learning methods we employ a 10-fold cross-validation strategy with a participant-wise split.
The different DL methods were trained for N number of epochs, where N was chosen based on the original papers.
The model weights from the final epoch were used to estimate the evaluation metric for each fold. The averaged metric for the 10 folds is reported in \autoref{tab: evaluation on datasets}.
The classical methods were directly evaluated on the whole datasets and the MAE is reported in \autoref{tab: evaluation on datasets}.

\begin{table}[!ht]
    \centering
    \caption{Performance (MAE) of methods on all datasets: Deep learning methods (top row) are evaluated using a 10-fold cross-validation strategy, with the overall MAE reported. Classical methods (bottom row) are evaluated on the entire dataset, and the MAE is reported. Standard Error (SE) is reported for all the methods.}
    \label{tab: evaluation on datasets}
    \begin{tabular}{lccccccc}
    \toprule
    \toprule
         \textbf{Methods}  & \multicolumn{2}{c}{ \textbf{PURE} } & \multicolumn{2}{c}{\textbf{COHFACE} }& \multicolumn{2}{c}{\textbf{CHILL}}  \\
         \midrule
         \midrule
        & MAE & SE & MAE & SE & MAE & SE \\  
        \cmidrule(lr){2-3} 
        \cmidrule(lr){4-5} 
        \cmidrule(lr){6-7} 
        DeepPhys    & \textbf{3.1} & 2.1 & 4.4  & 1.3     & 9.1  & 2.3 \\
        TS-CAN      & 10.1  & 5.5       & 4.1   & 2.1     & 3.2  & 0.2 \\ 
        Physnet     & 4   & 1.9       & \textbf{1.6} &  0.4 & 4.1 & 1.3\\
        rPPGNet     & 8.8   & 4.5       & 3.4  &  1.8      & \textbf{2} & 0.4 \\
        \midrule
               & MAE & SE & MAE & SE & MAE & SE \\
        \cmidrule(lr){2-3} 
        \cmidrule(lr){4-5} 
        \cmidrule(lr){6-7} 
        GREEN  & 10.1 & 2.9          & \textbf{7.1} & 0.7     & 2.6       & 0.6 \\ 
        CHROM  & 8.9  & 2.1          & 10.2         & 0.6     & 1.8       & 0.6\\ 
        POS    & 7.5  & 2             & 11.8         & 0.7     & \textbf{1.1} & 0.1 \\
        ICA    & \textbf{4.9} &  2  & 7.4          & 0.6     & 5.4       & 0.9 \\ 
        Dummy  & 15.6 & 2.2 & 9.7  & 0.74  & 10.8  & 0.7  \\  
    \bottomrule
        \end{tabular}
\end{table}

\begin{figure}
    \centering
    \includegraphics[width=\linewidth]{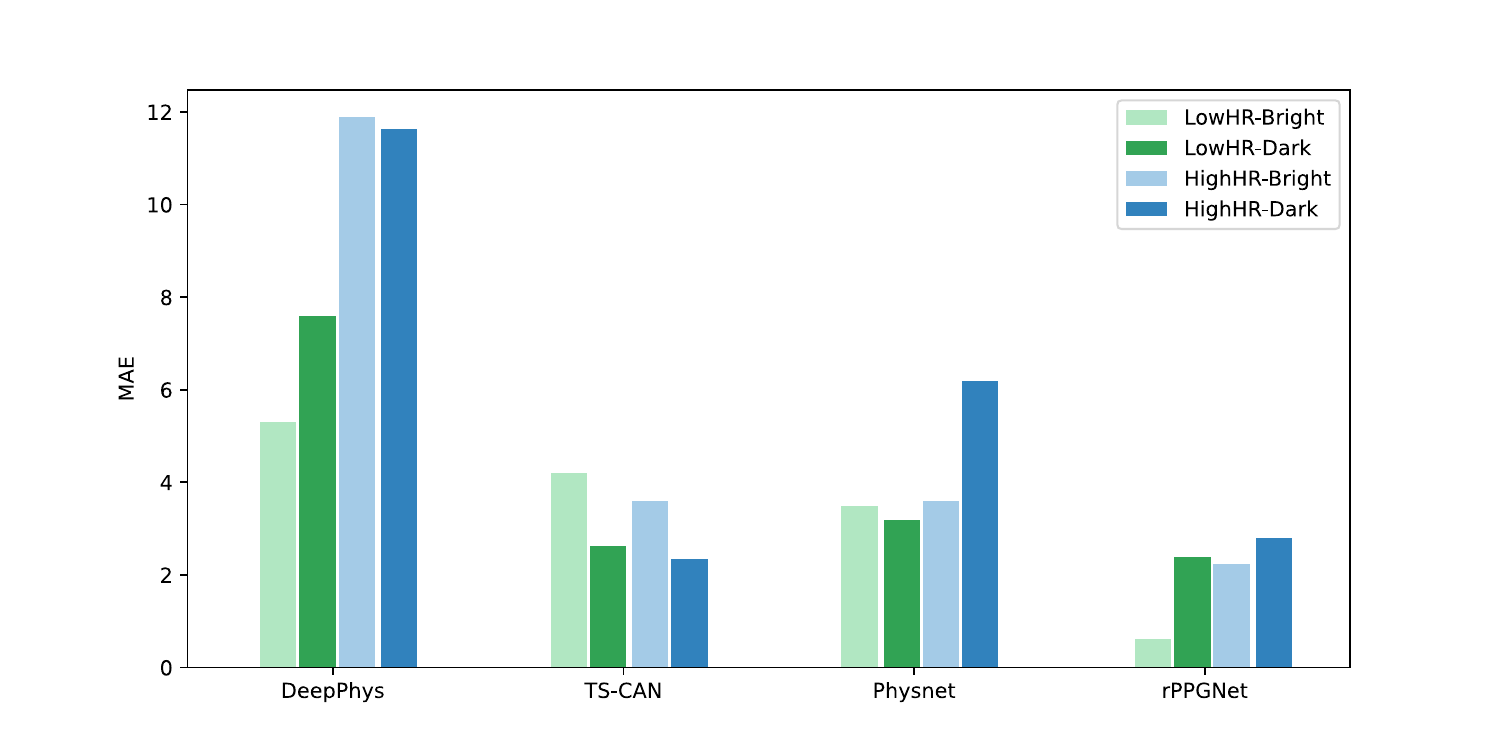}
    \caption{Performance (MAE) of DL on the different scenarios of the CHILL dataset}
    \label{fig: Performance (MAE) of DL methods on different scenarios of the CHILL}
\end{figure}
Looking first at the deep learning methods, we see that no model was consistently the best across all datasets.
DeepPhys \cite{chen2018deepphys} performed best on the PURE dataset \cite{stricker2014non}, while Physnet \cite{yu2019remote} and rPPGNet \cite{yu2019compress} performed the best on COHFACE \cite{Heusch_ARXIV_2017} and CHILL, respectively.
Additionally, some models showed variation in performance across datasets. TS-CAN \cite{liu2021multitask} and rPPGnet \cite{yu2019compress} perform poorly on PURE, while DeepPhys \cite{chen2018deepphys} performs poorly on CHILL. Notably, Physnet \cite{yu2019remote} demonstrated the most consistent performance across all the datasets. 
For the classical methods, we see that ICA \cite{poh2010advancements} and GREEN \cite{verkruysse2008remote} have the lowest MAE for PURE \cite{stricker2014non} and COHFACE \cite{Heusch_ARXIV_2017}, respectively. However, they are outperformed by deep learning methods. But for the CHILL dataset, the classical methods outperform the deep learning methods, where POS \cite{wang2016algorithmic} achieved the lowest MAE.

\begin{table*}
    \centering
    \caption{Performance (MAE) of DL models (trained on public datasets) and classical methods evaluated on CHILL dataset}
    \label{table:dl methods trained on public evaluated on CHILL}
    \begin{tabular}{cccccccc}
    \toprule
    \toprule
         \textbf{Trained On}& \textbf{ Models}&  \textbf{LowHR-Bright}&  \textbf{ LowHR-Dark}&  \textbf{HighHR-Bright}&  \textbf{HighHR-Dark} & \textbf{ALL}\\
 \midrule
 \midrule

COHFACE & DeepPhys   & 0.58                              & 1.27                            & 4.51                             & 4.98           & 2.84                      \\
        & TS-CAN     & \textbf{0.44}                              & \textbf{0.60}                            & 1.87                             & \textbf{1.38}           & \textbf{1.07}                      \\
        & Physnet    & 6.69                              & 4.17                            & 15.90                            & 4.70           & 7.86                      \\
        & rPPGNet    & 14.47                              &  27.59                            & 9.24                      & 20.96           &  18.06                    \\
 \midrule
PURE & DeepPhys   & 0.50                              & 0.67                            & \textbf{1.61}                             & 2.72              & 1.38                  \\
     & TS-CAN      &  0.42                              & 0.94                            & 1.80                             & 1.44             & 1.15                  \\
     & Physnet    & 12                             & 9.99                           & 20.63                            & 19.67             & 15.58                 \\
     & rPPGNet    & 8.95                             & 19.06                           & 18.08                            & 14.90             & 15.25                 \\
\midrule
 -                & GREEN &   1.11  & 1.45 & 5.15 & 2.68 & 2.60 \\
                  & CHROM &   0.53  & 0.73 & 3.08 & 2.72 & 1.76 \\
                  & POS   &   0.50  & 0.63 & 1.68 & 1.52 & 1.09   \\
                  & ICA   &   5.97  & 1.76 & 9.11 & 4.68 & 5.38 \\
    & Dummy &   9.62  & 9.81 &  12.41 &  11.25  & 10.75 \\
 \bottomrule
    \end{tabular}
\end{table*}

We next examined the model performance on different scenarios of our dataset. 
We present scenario-specific results of the 10-fold cross-validation on the CHILL dataset in \autoref{fig: Performance (MAE) of DL methods on different scenarios of the CHILL}. 
TS-CAN \cite{liu2021multitask} excelled in the HighHR-Dark scenario, while rPPGNet \cite{yu2019compress} outperformed other models in the remaining conditions.
Under similar illumination conditions, DeepPhys \cite{chen2018deepphys}, Physnet \cite{yu2019remote}, and rPPGNet \cite{yu2019compress} exhibited better performance in LowHR scenarios compared to HighHR. Conversely, TS-CAN has slightly better performance in HighHR scenarios.
When comparing LowHR scenarios (Bright and Dark) with their corresponding HighHR counterparts, we observe that DeepPhys \cite{chen2018deepphys}, Physnet \cite{yu2019remote}, and rPPGNet \cite{yu2019compress} exhibit better better performance in LowHR scenarios compared to HighHR. Conversely, TS-CAN \cite{liu2021multitask} has better performance in HighHR scenarios. 
However, the impact of illumination on performance varies across models. While rPPGNet \cite{yu2019compress} exhibits a decrease in performance under dark conditions, and TS-CAN \cite{liu2021multitask} demonstrate better performance in dark scenarios.
ICA \cite{poh2010advancements} and GREEN \cite{verkruysse2008remote} have the lowest MAE for PURE and COHFACE, respectively.

\subsection{Generalization of Methods to CHILL Dataset}
\label{subsec: generalization of dl models trained on public datasets to CHILL}
We aim to assess the generalization ability of deep learning models trained on publicly available datasets to our novel dataset. To accomplish this, we train the deep learning methods on the entire public datasets while reserving the complete CHILL dataset for testing. As a comparison baseline, we also present a dummy estimator that always predicts the mean heart rate.

The results from the experiment are summarized in \autoref{table:dl methods trained on public evaluated on CHILL}. 
Additionally, we also present the scenarios specific evaluations of the classical methods on the CHILL dataset in \autoref{table:dl methods trained on public evaluated on CHILL}.
Two methods stand out for their performance: TS-CAN and POS, with a difference of 0.02 BPM in overall MAE.
Among the deep learning models pre-trained on COHFACE, DeepPhys, and TS-CAN exhibit higher performance in LowHR scenarios compared to the HighHR. 
Furthermore, their performance in the LowHR setting decreases in the dark setting. 
Conversely, Physnet and rPPGnet perform better in dark scenarios compared to that of bright scenarios in LowHR. 
All methods pre-trained on PURE, except rPPGNet, exhibit better performance in LowHR scenarios compared to HighHR. 
Notably, TS-CAN, the best-performing model among them, shows a decrease in performance under dark conditions for both LowHR and HighHR scenarios.

Overall, it can be seen that the DeepPhys and TS-CAN have consistently achieved lower MAEs regardless of the pertaining dataset while the other methods have high MAEs. 
Furthermore, all methods pre-trained on COHFACE, except DeepPhys, have better performance compared to models pre-trained on PURE. 

Finally, examining the classical methods, we observe that POS and CHROM outperform the others.
These two methods exhibit a slight decrease in performance for LowHR scenarios under dark conditions, but an increase in performance for HighHR scenarios under dark conditions. ICA and GREEN perform better in dark scenarios compared to bright ones in both LowHR and HighHR.
\section{Discussion}
In this work, we evaluated the performance of rPPG estimation methods under challenging conditions.
To this end, we collect a novel dataset that includes scenarios such as low illumination and high heart rate.
This dataset is also made available to other researchers. 
Our evaluations revealed two key findings regarding rPPG performance under such challenging conditions. First, rPPG methods, including both classical and deep learning approaches, generally exhibit lower performance in high heart rate conditions. Interestingly, the impact of illumination on deep learning methods varied, with specific performance changes depending on the chosen method and the training dataset. Second, classical methods, which often perform poorly on publicly available datasets, surprisingly outperformed deep learning methods on our dataset.

Firstly, upon examining the performance of classical methods across the entire CHILL dataset (\autoref{table:dl methods trained on public evaluated on CHILL}), it becomes apparent that these methods are not greatly impacted by low-light conditions (LowHR-Dark - HighHR-Dark). However, it is notable that more intricate techniques like CHROM and POS, which are constructed based on the skin reflectance model \cite{wang2016algorithmic}, outshine simpler approaches such as GREEN and ICA.
Nevertheless, these methods exhibit poor performance on existing public datasets (\autoref{tab: evaluation on datasets}). 

The evaluation in \ref{subsec: training on datasets} confirms previous research, demonstrating that deep learning methods outperform the classical methods on existing public datasets. 
However, no single method emerges as the overall best performer across all datasets.  
Their performance on the different settings of the CHILL dataset revealed no clear trend based on illumination for any of these methods. However, all methods except TS-CAN exhibited a decrease in performance in the presence of high heart rate.
Upon further investigation, we discovered that TS-CAN performed poorly on a specific fold (consisting of 5 participants) of the low heart rate settings. When we excluded this particular fold, we observed that TS-CAN exhibited a performance pattern similar to the other models, with a drop in performance under high heart rate scenarios.

Our evaluations on the generalizability of rPPG methods, as outlined in \autoref{table:dl methods trained on public evaluated on CHILL} highlight the strong performance of DeepPhys and TS-CAN on our dataset. 
These methods exhibit minimal performance changes in response to changes in illumination.
In this regard, our findings regarding DeepPhys diverge from those of Yang et al. \cite{yang2022assessment}. Their study showed a drastic drop in the performance of DeepPhys for low illumination settings, which is not the case in our experiments. Furthermore, the performance of the DeepPhys is close to that of POS, which had the best performance on CHILL dataset.
However, other deep learning methods in our study experienced a noticeable decline in performance, aligning with the observations of Yang et al. \cite{yang2022assessment}. 

We also observed that TS-CAN \cite{liu2021multitask} outperformed classical methods in the low heart rate (LowHR) setting, while POS emerged as the overall best performer on the CHILL dataset. This further highlights the ability of deep learning methods to adapt to illumination variations, while also revealing their vulnerability to high heart rates (HighHR).
Throughout our analysis, it is evident that no single method emerges as universally superior across all scenarios and datasets. Researchers should take this into account when applying these methods to other downstream tasks. It's crucial to conduct a thorough evaluation to determine the method that aligns best with their specific application.
To this end, our collected dataset which consists of elevated heart rates and low illumination scenarios can be used by researchers to enrich publicly available datasets. 
The feasibility of this enrichment will be investigated in future studies.
\section{Conclusion}
In this work, we investigated the performance of four classical and four deep learning-based rPPG methods under challenging conditions. We evaluated these methods on two publicly available datasets and our novel dataset specifically designed to include elevated heart rates and low illumination scenarios.
Our evaluations revealed that both classical and deep learning methods showed decreased performance in high heart rate scenarios. Surprisingly, classical methods outperformed deep learning on our novel dataset. Deep learning method performance under illumination variations depended on the specific method and its training data.

{\small
\bibliographystyle{ieee_fullname}
\bibliography{main}
}

\end{document}